\documentclass[conference]{IEEETran}
\usepackage{float}
\usepackage[utf8]{inputenc} 
\usepackage[T1]{fontenc}    
\usepackage{hyperref}       
\usepackage{url}            
\usepackage{booktabs}       
\usepackage{amsfonts}       
\usepackage{nicefrac}       
\usepackage{microtype}      
\usepackage{xcolor}         
\usepackage{amsmath,amssymb}
\usepackage{wrapfig}
\usepackage{stfloats}
\usepackage{graphicx}
\usepackage{caption}
\usepackage{subcaption}
\usepackage{bm}

\usepackage{listings}
\NewDocumentCommand{\codeword}{v}{%
\texttt{\textcolor{blue}{#1}}%
}

\newcommand{\Rb}{\mathbb{R}}

\title{SigGate: Enhancing Recurrent Neural Networks with Signature-Based Gating Mechanisms}

\author{%
  \IEEEauthorblockN{%
    Rémi Genet\IEEEauthorrefmark{1}\textsuperscript{\textsection} and
    Hugo Inzirillo\IEEEauthorrefmark{2}\textsuperscript{\textsection}
  }%
  
  \IEEEauthorblockA{\IEEEauthorrefmark{1} DRM, Université Paris Dauphine - PSL}%
    \IEEEauthorblockA{\IEEEauthorrefmark{2} CREST-ENSAE, Institut Polytechnique de Paris}%
}

\begin{document}
\thispagestyle{plain}
\pagestyle{plain}
\maketitle
\begingroup\renewcommand\thefootnote{\textsection}
\footnotetext{These authors contributed equally.}
\endgroup

\begin{abstract}
In this paper, we propose a novel approach that enhances recurrent neural networks (RNNs) by incorporating path signatures into their gating mechanisms. Our method modifies both Long Short-Term Memory (LSTM) and Gated Recurrent Unit (GRU) architectures by replacing their forget and reset gates, respectively, with learnable path signatures. These signatures, which capture the geometric features of the entire path history, provide a richer context for controlling information flow through the network's memory. This modification allows the networks to make memory decisions based on the full historical context rather than just the current input and state. Through experimental studies, we demonstrate that our Signature-LSTM (SigLSTM) and Signature-GRU (SigGRU) models outperform their traditional counterparts across various sequential learning tasks. By leveraging path signatures in recurrent architectures, this method offers new opportunities to enhance performance in time series analysis and forecasting applications.
\end{abstract}
\thispagestyle{plain}
\pagestyle{plain}
\section{Introduction}
The modeling of sequential data is crucial in machine learning application, particularly in time series analysis \cite{shumway2000time} and forecasting. Understanding temporal dependencies is fundamental to make accurate prediction.  Traditional methods such as autoregressive models \cite{box2015time} and hidden Markov models \cite{Mar13} often struggle to capture these long term dependencies in highly complex nonlinear sequences \cite{box2015time}. With the advent of deep learning \cite{lecun2015deep}, Recurrent Neural Networks (RNNs) \cite{medsker1999recurrent} have emerged as a powerful tool for modeling sequential data, demonstrating remarkable success in tasks ranging from financial forecasting \cite{lim2021time,genet2024tkan,lim2021temporal,genet2024temporal}, to speech recognition and healthcare analytics  Among RNN variants,  Long Short-Term Memory (LSTM) \cite{hochreiter1997long} and Gated Recurrent Units (GRU) \cite{cho2014learning} are particularly successful due to their ability to capture long-term dependencies. These architectures rely on gating mechanisms to control the flow of information through their memory cells, traditionally using the current input and previous state to make these decisions. However, this local view might limit their ability to capture broader temporal patterns and relationships in the data.
Path signatures, first introduced by Chen \cite{chen1958integration}, offer a powerful mathematical framework for encoding the essential features of sequential data through iterated integrals. Recent works have demonstrated the utility of signatures in various machine learning applications \cite{chevyrev2016primer,fermanian2021embedding}, particularly in time series analysis \cite{gyurko2013extracting,dyer2021deep}. The key insight is that path signatures can capture important geometric and temporal patterns in the data, providing a rich representation of the entire history of a sequence. In our recent work \cite{inzirillo2024sigkan} we have shown that incorporating path signatures as weighting mechanisms within neural networks can significantly improve their performance on sequential tasks.
Building upon these foundations, we propose a novel modification to LSTM and GRU architectures that leverages path signatures to enhance their gating mechanisms. In our approach, we replace the forget gate in LSTM and the reset gate in GRU with signature-based gates that compute their values using the signature of the input sequence up to the current time step. This modification allows these gates to make decisions based on the entire history of the input sequence, rather than just the current input and state. The signature stream provides a natural way to capture the evolving dynamics of the input sequence, making it particularly well-suited for controlling memory retention and reset decisions in recurrent architectures.
Our work demonstrates that integrating path signatures directly into the gating mechanisms of recurrent networks leads to improved performance across various sequential learning tasks. The resulting  Signature-LSTM (SigLSTM) and  Signature-GRU (SigGRU) architectures show an enhanced ability to capture long-term dependencies while maintaining the computational efficiency of their traditional counterparts. These improvements are particularly notable in tasks requiring an understanding of complex temporal patterns and long-range interactions in the data.
The implementation of the SigLSTM and SigGRU are available as an open-source Python package \texttt{sig\_rnn} on PyPI, and the complete source code can be found at \href{https://github.com/remigenet/sig_rnn/}{sig\_rnn}. This repository includes all the code necessary to reproduce our experiments.

\section{Related Work}
Recurrent Neural Networks (RNNs), particularly Long Short-Term Memory (LSTM) and Gated Recurrent Unit (GRU) architectures, have been widely employed in sequential learning tasks due to their ability to capture temporal dependencies. However, traditional gating mechanisms in these models primarily rely on local context using only the "current" input and "hidden state" for memory updates which can limit their effectiveness in modeling long-range dependencies \cite{gu2020improving}. Some researchers proposed refinements on the gating mechanisms to enhance gradient flow and learning \cite{lu2017simplified,gu2020improving,cheng2020refined}. At the same time, the concept of path signatures, derived from rough path theory, has emerged as a key mathematical tool for deciphering the geometric structure of sequential data \cite{lyons1998differential}. These signatures have been successfully adapted in neural architectures, enhancing their ability to process complex temporal patterns. In particular, \cite{sabate2020solving} showed that integrating these signatures into LSTM networks optimized the solution of path-dependent partial differential equations, with applications in finance. In the same vein, \cite{moreno2024rough} recently proposed Rough Transformers, models exploiting path signatures to improve the analysis of continuous-time data. Recent studies have also investigated the application of path signatures in RNN-based models. \cite{gu2020improving} introduced alternative gating mechanisms to address saturation issues in traditional RNNs, while path-based feature extraction has been utilized in areas such as personalized user modeling (e.g., Sequential Path Signature Networks) and path classification tasks. These signatures have been also leveraged in classification tasks in finance to build a novel approach to portfolio construction for digital assets \cite{inzirillo2024clustering}. These advancements indicate that incorporating path signatures into deep learning frameworks can enhance the representation of both short-term and long-term sequences.  Building upon these advances, our work proposes a novel SigLSTM and SigGRU, where we replace the forget and reset gates, respectively, with learnable path signatures. This allows for richer contextual modeling by incorporating the full historical trajectory into memory updates. Our approach extends prior work by leveraging path signatures in a fully learnable manner within standard recurrent architectures, improving performance across diverse sequence modeling and time series forecasting tasks.

\section{Model Architecture}
At their core, Recurrent Neural Networks with gated architectures aim to solve the vanishing gradient problem encountered in simple RNNs by introducing a memory cell. This memory cell maintains information over time and is controlled by different gates that determine what information should be kept, forgotten, or updated. The two most prominent architectures implementing this concept are the Long Short-Term Memory (LSTM) and the Gated Recurrent Unit (GRU).
\subsubsection{Long Short-Term Memory}
The LSTM architecture maintains a cell state $c_t$ that acts as a memory conveyor belt, with gates controlling the flow of information. At each time step $t$, given an input $x_t$ and the previous hidden state $h_{t-1}$, the LSTM computes:
\begin{equation}\tag{Input Gate}
    i_t = \sigma(W_i x_t + U_i h_{t-1} + b_i),
\end{equation}
\begin{equation}\tag{Forget Gate}
    f_t = \sigma(W_f x_t + U_f h_{t-1} + b_f),
\end{equation}
\begin{equation}\tag{Cell Gate}
    \tilde{c}_t = \tanh(W_c x_t + U_c h_{t-1} + b_c),
\end{equation}
\begin{equation}\tag{Output Gate}
    o_t = \sigma(W_o x_t + U_o h_{t-1} + b_o),
\end{equation}
The cell state and hidden state are then updated as:
\begin{equation}\tag{Cell Update}
    c_t = f_t \odot c_{t-1} + i_t \odot \tilde{c}_t,
\end{equation}
\begin{equation}\tag{Hidden State}
    h_t = o_t \odot \tanh(c_t).
\end{equation}
\subsubsection{Gated Recurrent Unit}
The GRU simplifies the LSTM architecture by combining the forget and input gates into a single update gate and merging the cell state and hidden state. Given an input $x_t$ and previous hidden state $h_{t-1}$, the GRU computes:
\begin{equation}\tag{Update Gate}
    z_t = \sigma(W_z x_t + U_z h_{t-1} + b_z)
\end{equation}
\begin{equation}\tag{Reset Gate}
    r_t = \sigma(W_r x_t + U_r h_{t-1} + b_r)
\end{equation}
\begin{equation}\tag{Candidate State}
    \tilde{h}_t = \tanh(W_h x_t + U_h(r_t \odot h_{t-1}) + b_h).
\end{equation}
The hidden state is then updated as:
\begin{equation}\tag{Hidden Update}
    h_t = (1-z_t) \odot h_{t-1} + z_t \odot \tilde{h}_t.
\end{equation}
In both architectures, $\sigma$ represents the sigmoid activation function, $\tanh$ is the hyperbolic tangent activation function, $\odot$ denotes element-wise multiplication, and $W_*, U_*, b_*$ are learnable parameters.

\subsection{Path Signatures}
Path signatures provide a systematic way to encode sequential data by computing iterated integrals along a path. For a path $X: [0,T] \rightarrow \Rb^d$, the signature captures information about both the path's values and their patterns of variation.
Given an N-dimensional path $(X_t)_{t \in [0,T]}$ where $X_t = (X_{1,t},...,X_{N,t})$, the signature is constructed through successive levels of iterated integrals. The first level consists of the increments of each coordinate path over the interval $[0,t]$:
\begin{equation}\tag{Level 1}
    S(X)^n_{0,t} = \int_0^t dX^n_s, \quad n \in \{1,...,N\}.
\end{equation}
The second level captures pairwise interactions between coordinates through double iterated integrals:
\begin{equation}\tag{Level 2}
    S(X)^{n,m}_{0,t} = \int_0^t S(X)^n_{0,s} dX^m_s, \quad n,m \in \{1,...,N\},
\end{equation}
This pattern continues for higher levels, with the k-th level signature terms defined recursively as:
\begin{equation}\tag{Level k}
    S(X)^{i_1,...,i_k}_{0,t} = \int_0^t S(X)^{i_1,...,i_{k-1}}_{0,s} dX^{i_k}_s,
\end{equation}
where $i_1,...,i_k \in \{1,...,N\}$. The complete signature is the collection of all these terms:
\begin{equation}\tag{Full Signature}
    \begin{aligned}
        S(X)_{0,T} = (1, S(X)^1_{0,T},...,S(X)^N_{0,T}, \\
        S(X)^{1,1}_{0,T},...,S(X)^{N,N}_{0,T}, \\
        S(X)^{1,1,1}_{0,T},...)
    \end{aligned}
\end{equation}
In practice, we truncate this infinite series at a specified depth $M$, resulting in a finite-dimensional vector. The dimension of the truncated signature grows as $\sum_{k=1}^M N^k$, where $N$ is the dimension of the input path. For computational efficiency and to maintain a reasonable number of parameters, we typically use $M=2$ or $M=4$ in our models.
To normalize the signature values across different sequence lengths, we apply time normalization:
\begin{equation}\tag{Time Normalization}
    \hat{S}(X)_{0,t} = \frac{S(X)_{0,t}}{t}.
\end{equation}
This normalization helps ensure that the signature terms remain comparable across different time scales and sequence lengths, which is crucial for their use in neural network architectures.
\subsection{Integration of Path Signatures in Gating Mechanisms}
Our approach enhances traditional RNN architectures by replacing specific gates with signature-based computations. Before computing the signatures, we first reduce the dimensionality of the input sequence through a learnable linear transformation:
\begin{equation}\tag{Input Preprocessing}
    \tilde{X}_t = W_{sig} X_t
\end{equation}
where $W_{sig} \in \Rb^{5 \times d}$ projects the d-dimensional input to a 5-dimensional space. This preprocessing step is crucial for computational efficiency, as the signature dimension grows exponentially with the input dimension. For instance, with a hidden state of size 100, a direct signature computation of degree 2 would result in a signature of dimension 10,000, making the model computationally intractable.

In the SigLSTM, we replace the traditional forget gate computation with a signature-based version:

\begin{equation}\tag{Signature Forget Gate}
    f_t = \sigma(W_f S(\tilde{X})_{0,t} + b_f)
\end{equation}
where $S(\tilde{X})_{0,t}$ represents the time-normalized signature of the preprocessed input sequence up to time t, and $W_f$ is now a matrix that maps from the signature space to the hidden state dimension.
Similarly, for the SigGRU, we modify the reset gate computation:
\begin{equation}\tag{Signature Reset Gate}
    r_t = \sigma(W_r S(\tilde{X})_{0,t} + b_r)
\end{equation}
The other gates and computations in both architectures remain unchanged from their traditional counterparts. This targeted modification allows us to leverage the rich temporal information captured by signatures specifically for memory control decisions, while maintaining the overall structure and efficiency of the original architectures.The signature stream provides these gates with a comprehensive view of the input history, allowing them to make more informed decisions about memory retention. The path signature captures not just the current state of the input, but also how it has evolved over time, including important geometric features and temporal patterns that might be relevant for determining what information should be maintained or discarded from the memory cell.

\subsection{Implementation Details}
The implementation of both SigLSTM and SigGRU leverages the multi-backend capabilities of Keras, making these architectures compatible with TensorFlow, JAX, and PyTorch backends. The signature computations utilize the keras-sig package, which provides efficient GPU-accelerated signature calculations within Keras models. By enabling the stream option in keras-sig, we obtain the complete sequence of iterated integrals up to time t for each timestep, which is essential for our sequential processing. For computational efficiency, we extensively use Einstein summation operations (einsum) to batch-compute transformations across the entire sequence. For instance, in both architectures, we compute the input transformations for all timesteps simultaneously. Let $X \in \mathbb{R}^{B \times T \times I} $ be the input tensor, where $B $ represents the batch size, $T $ represents the number of time steps, and $I $ represents the input feature dimension. Similarly, let $K \in \mathbb{R}^{I \times J} $ denote the kernel matrix (or weight matrix), where $I $ is the input feature dimension (matching the last dimension of $X $), and $J $ is the output feature dimension, which defines the dimensionality of the transformation applied to the input tensor.The output $Y \in \mathbb{R}^{B \times T \times J} $, is computed as:
\begin{equation}
    \begin{split}
        Y_{b,t,j} &= \sum_{i=1}^{I} X_{b,t,i} \cdot K_{i,j}, \\ & \forall b \in \{1, \dots, B\}, \ t \in \{1, \dots, T\}, \ j \in \{1, \dots, J\}.
    \end{split}
\end{equation}
This equation specifies that for each batch $b $, each time step $t $, and each output feature $j $, the corresponding output $Y_{b,t,j} $ is computed by summing over the product of $X_{b,t,i} $ (the input feature at batch $b$, time step $t $, and feature $i $) and $K_{i,j}$ (the weight from input feature $i $ to output feature $j $). This operation transforms the input features $X $ into a new feature space defined by $K $, while preserving the batch and time dimensions.This operation efficiently combines the input sequence with the weight matrices in a single operation, avoiding the need for explicit loops. When using the JAX backend, we implement the sequential processing using the scan operation, which offers significant advantages over traditional loops. The scan operation allows explicit control over the unroll level of the computation, making the compilation process more manageable. This implementation strategy results in models that are both computationally efficient and practical for deployment across different backend environments, while maintaining the theoretical advantages of signature-based gating mechanisms.

\section{Learning Tasks}
We evaluate our signature-based RNN architectures on two distinct forecasting tasks using cryptocurrency market data from the Binance exchange, spanning from January 1, 2020, to December 31, 2022. Our dataset focuses on USDT markets, which represent the most actively traded pairs on the exchange. For the learning task we used the package Keras Sig \cite{genet2025keras}.
\subsection{Task Definitions}
The first task involves predicting trading volumes, utilizing hourly notional amounts traded across multiple assets (BTC, ETH, ADA, XMR, EOS, MATIC, TRX, FTM, BNB, XLM, ENJ, CHZ, BUSD, ATOM, LINK, ETC, XRP, BCH, and LTC) to forecast BTC volume. This multivariate approach allows the models to capture potential cross-market dependencies and leading indicators. The second task focuses on predicting absolute returns of BTC prices, using the same multivariate input structure. This modification from the original univariate setup in prior work enables us to investigate whether cross-market information can enhance volatility forecasting accuracy, while absolute returns provide a direct measure of price variability. 

\subsection{Data Preprocessing}
For volume prediction, we use a two-stage scaling approach. First, we divide each series by its two-week moving median, shifted forward by the prediction horizon to prevent look-ahead bias. This transformation helps achieve stationarity across time. Subsequently, we apply MinMax scaling per asset to constrain values within [0,1], using only training data to compute scaling parameters. For absolute returns prediction, we maintain the same multivariate structure but scale the data using the maximum value from the training set to ensure all values fall within [0,1], with the possibility of test set values exceeding this range.

\subsection{Model Variants and Training Setup}
All models are implemented with a consistent hidden state size of 100 units across all recurrent layers. We evaluate several architectural variants:
\subsubsection{Signature-Based Models}
Our signature-based architectures explore different configurations while maintaining the base hidden size:
\begin{itemize}
    \item Varying signature depths (2, 3, and 4) with consistent architecture depth
    \item Different input projection dimensions (default 5 and expanded 10)
    \item Architectural variations (2-layer, 3-layer, and flattened output versions)
\end{itemize}

\subsubsection{Baseline Models}
Standard LSTM and GRU models are implemented with matching architectures and the same hidden size of 100 units, ensuring a fair comparison with our signature-based versions.

\subsection{Training Protocol}
The training process was configured with a batch size of 128 and a maximum of 1000 epochs. Early stopping was employed to prevent overfitting, with a patience of 10 epochs and a minimum delta of 0.00001 to determine significant improvement. Additionally, a learning rate reduction on plateau mechanism was implemented, reducing the learning rate by a factor of 0.25 after 5 epochs of no improvement, with a minimum learning rate set at 0.000025. The training data was split with 20\% reserved for validation, ensuring robust performance evaluation during training.
The sequence length for all models is set to the maximum between 45 and 5 times the prediction horizon, providing sufficient historical context while maintaining computational efficiency. For volume prediction, we evaluate horizons of 1, 9, and 15 steps ahead, while for absolute returns prediction, we focus on 1 and 6 steps ahead. Each model configuration is trained and evaluated five times to assess stability and statistical significance of the results.

\section{Results Analysis}

Let us first examine the performance on the volatility prediction task. Table \ref{tab:vol_results_with_std} presents the $R^2$ scores across different model architectures.
\begin{table}[htbp]
\centering
\caption{Average and Standard Deviation of \(R^2\) Scores for Absolute Returns Prediction}
\scalebox{0.9}{
\begin{tabular}{lcccc}
\toprule
\textbf{Model} & \multicolumn{2}{c}{\textbf{1-step ahead}} & \multicolumn{2}{c}{\textbf{6-step ahead}} \\
& Mean & Std & Mean & Std \\
\midrule
\multicolumn{5}{l}{\textit{Two-Layer LSTM Models}} \\
LSTM & 0.1620 & 0.0017 & 0.1257 & 0.0075 \\
SignatureLSTM-2-2 & 0.1597 & 0.0099 & \underline{0.1313} & 0.0031 \\
SignatureLSTM-3-2 & \underline{\textbf{0.1642}} & 0.0043 & 0.1312 & 0.0084 \\
SignatureLSTM-3-3 & 0.1544 & 0.0103 & 0.1312 & 0.0068 \\
SignatureLSTM-4-4 & 0.1625 & 0.0033 & 0.1303 & 0.0048 \\
\midrule
\multicolumn{5}{l}{\textit{Two-Layer GRU Models}} \\
GRU & \underline{0.1604} & 0.0033 & \underline{0.1263} & 0.0070 \\
SignatureGRU-2-2 & 0.1531 & 0.0072 & 0.1135 & 0.0077 \\
SignatureGRU-3-2 & 0.1503 & 0.0097 & 0.1135 & 0.0117 \\
SignatureGRU-3-3 & 0.1519 & 0.0107 & 0.1151 & 0.0038 \\
SignatureGRU-4-4 & 0.1589 & 0.0042 & 0.1222 & 0.0027 \\
\midrule
\multicolumn{5}{l}{\textit{Three-Layer Models}} \\
LSTM-3 & 0.1533 & 0.0106 & 0.1248 & 0.0047 \\
SignatureLSTM-3-3-3 & \underline{0.1637} & 0.0029 & \underline{\textbf{0.1324}} & 0.0057 \\
GRU-3 & 0.1564 & 0.0071 & 0.1190 & 0.0155 \\
SignatureGRU-3-3-3 & 0.1502 & 0.0087 & 0.1143 & 0.0067 \\
\midrule
\multicolumn{5}{l}{\textit{Flattened Output Models}} \\
LSTM & 0.1551 & 0.0036 & 0.1234 & 0.0049 \\
SignatureLSTM-3-3 & 0.1515 & 0.0085 & 0.1175 & 0.0069 \\
GRU & 0.1429 & 0.0131 & 0.1156 & 0.0022 \\
SignatureGRU-3-3 & 0.1444 & 0.0054 & 0.1119 & 0.0045 \\
\midrule
\multicolumn{5}{l}{\textit{Extended Signature Input Models}} \\
SignatureLSTM-2-2 & \underline{0.1631} & 0.0057 & \underline{0.1298} & 0.0054 \\
SignatureGRU-2-2 & 0.1530 & 0.0055 & 0.1111 & 0.0039 \\
\bottomrule
\end{tabular}
}
\label{tab:vol_results_with_std}
\end{table}
For short-term predictions (1-step ahead), we observe that LSTM-based models generally outperform their GRU counterparts, with traditional LSTM achieving an $R^2$ of 0.1620. However, signature-enhanced variants demonstrate superior performance, with SignatureLSTM-3-2 reaching the highest $R^2$ of 0.1642. This improvement, while modest, is consistent across multiple runs as indicated by the low standard deviation of 0.0043. For medium-term predictions (6-step ahead), the performance dynamics shift significantly. Here, signature-based architectures show a more substantial advantage, with SignatureLSTM-3-3-3 achieving an $R^2$ of 0.1324. This represents a marked improvement over traditional models, where performance typically degrades more rapidly with prediction horizon. The computational requirements for these models, shown in Table \ref{tab:training_times}, reveal important practical considerations.
\begin{table}[htbp]
\centering
\caption{Average Training Time (seconds) for Model Variants}
\scalebox{0.9}{
\begin{tabular}{lcc}
\toprule
\textbf{Model} & \textbf{1-step ahead} & \textbf{6-step ahead} \\
\midrule
\multicolumn{3}{l}{\textit{Two-Layer LSTM Models}} \\
LSTM & 14.88 & 14.27 \\
SignatureLSTM-2-2 & 80.62 & 87.69 \\
SignatureLSTM-3-2 & 86.94 & 93.13 \\
SignatureLSTM-3-3 & 86.83 & 91.79 \\
SignatureLSTM-4-4 & 93.15 & 98.96 \\
\midrule
\multicolumn{3}{l}{\textit{Two-Layer GRU Models}} \\
GRU & 17.03 & 15.81 \\
SignatureGRU-2-2 & 39.00 & 39.66 \\
SignatureGRU-3-2 & 43.15 & 42.79 \\
SignatureGRU-3-3 & 42.20 & 42.11 \\
SignatureGRU-4-4 & 48.74 & 46.76 \\
\midrule
\multicolumn{3}{l}{\textit{Three-Layer Models}} \\
LSTM & 21.06 & 20.58 \\
SignatureLSTM-3-3-3 & 122.06 & 130.08 \\
GRU & 24.41 & 23.20 \\
SignatureGRU-3-3-3 & 55.36 & 57.53 \\
\midrule
\multicolumn{3}{l}{\textit{Flattened Output Models}} \\
LSTM & 14.42 & 14.82 \\
SignatureLSTM-3-3 & 116.28 & 95.98 \\
GRU & 16.61 & 17.96 \\
SignatureGRU-3-3 & 42.77 & 40.37 \\
\midrule
\multicolumn{3}{l}{\textit{Extended Signature Input Models}} \\
SignatureLSTM-2-2 & 85.12 & 92.14 \\
SignatureGRU-2-2 & 42.39 & 39.38 \\
\bottomrule
\end{tabular}
}
\label{tab:training_times}
\end{table}
While signature-based models demonstrate superior predictive performance, they come with increased computational costs. The training time for SignatureLSTM variants is typically 5-6 times longer than traditional LSTMs, though SignatureGRU models show a more moderate increase of 2-3 times. This computational overhead must be weighed against the performance benefits when considering deployment scenarios.
Turning to the volume prediction task, Table \ref{tab:volume_results_with_std} presents a notably different performance pattern.
\begin{table}[htbp]
\centering
\caption{Average and Standard Deviation of \(R^2\) Scores for Volume Prediction}
\scalebox{0.8}{
\begin{tabular}{lcccccc}
\toprule
\textbf{Model} & \multicolumn{2}{c}{\textbf{1-step ahead}} & \multicolumn{2}{c}{\textbf{9-step ahead}} & \multicolumn{2}{c}{\textbf{15-step ahead}} \\
& Mean & Std & Mean & Std & Mean & Std \\
\midrule
\multicolumn{7}{l}{\textit{Two-Layer LSTM Models}} \\
LSTM & 0.3847 & 0.0061 & -0.0934 & 0.1288 & -0.1718 & 0.1369 \\
SignatureLSTM-2-2 & 0.3931 & 0.0171 & -0.1346 & 0.2016 & -0.1072 & 0.0940 \\
SignatureLSTM-3-2 & \underline{0.4010} & 0.0059 & \underline{-0.0399} & 0.0521 & \underline{-0.0396} & 0.1167 \\
SignatureLSTM-3-3 & 0.3996 & 0.0136 & -0.1359 & 0.0828 & -0.1486 & 0.1463 \\
SignatureLSTM-4-4 & 0.3956 & 0.0101 & -0.1228 & 0.0944 & -0.0779 & 0.0726 \\
\midrule
\multicolumn{7}{l}{\textit{Two-Layer GRU Models}} \\
GRU & 0.3972 & 0.0018 & 0.0949 & 0.0224 & 0.0807 & 0.0411 \\
SignatureGRU-2-2 & 0.3995 & 0.0070 & 0.1183 & 0.0098 & 0.0944 & 0.0045 \\
SignatureGRU-3-2 & 0.4029 & 0.0032 & 0.1164 & 0.0077 & 0.0833 & 0.0080 \\
SignatureGRU-3-3 & \underline{\textbf{0.4051}} & 0.0019 & \underline{\textbf{0.1210}} & 0.0078 & \underline{\textbf{0.0963}} & 0.0062 \\
SignatureGRU-4-4 & 0.4042 & 0.0048 & 0.1196 & 0.0041 & 0.0891 & 0.0076 \\
\midrule
\multicolumn{7}{l}{\textit{Three-Layer Models}} \\
LSTM & 0.3700 & 0.0212 & -0.0662 & 0.1228 & -0.1644 & 0.0467 \\
SignatureLSTM-3-3-3 & \underline{0.4025} & 0.0036 & -0.2870 & 0.1957 & -0.1309 & 0.0725 \\
GRU & 0.3840 & 0.0079 & 0.0614 & 0.0536 & 0.0253 & 0.0634 \\
SignatureGRU-3-3-3 & 0.3941 & 0.0045 & \underline{0.1185} & 0.0080 & \underline{0.0800} & 0.0065 \\
\midrule
\multicolumn{7}{l}{\textit{Flattened Output Models}} \\
LSTM & 0.3861 & 0.0071 & -0.2347 & 0.2266 & -0.3657 & 0.4826 \\
SignatureLSTM-3-3 & \underline{0.4038} & 0.0028 & 0.0775 & 0.0368 & -0.0027 & 0.0469 \\
GRU & 0.3878 & 0.0026 & 0.0891 & 0.0107 & \underline{0.0759} & 0.0041 \\
SignatureGRU-3-3 & 0.3933 & 0.0056 & \underline{0.1136} & 0.0088 & 0.0669 & 0.0123 \\
\midrule
\multicolumn{7}{l}{\textit{Extended Signature Input Models}} \\
SignatureLSTM-2-2 & 0.3898 & 0.0132 & -0.1497 & 0.1205 & -0.1256 & 0.0867 \\
SignatureGRU-2-2 & \underline{0.4041} & 0.0037 & \underline{0.1207} & 0.0114 & \underline{0.0900} & 0.0037 \\
\bottomrule
\end{tabular}
}
\label{tab:volume_results_with_std}
\end{table}
In contrast to the volatility task, GRU-based architectures consistently outperform LSTM variants for volume prediction, particularly at longer horizons. Standard LSTM models struggle with nine and fifteen-step predictions, often yielding negative $R^2$ values, while GRU maintains positive performance. The SignatureGRU-3-3 achieves the best results across all horizons ($R^2$ of 0.4051, 0.1210, and 0.0963 for one, nine, and fifteen steps respectively), demonstrating remarkable stability as indicated by low standard deviations.The timing results for volume prediction, presented in Table \ref{tab:volume_training_times}, follow similar patterns to the volatility task.
\begin{table}[htbp]
\centering
\caption{Average Training Time (seconds) for Volume Prediction}
\scalebox{0.9}{
\begin{tabular}{lccc}
\toprule
\textbf{Model} & \textbf{1-step ahead} & \textbf{9-step ahead} & \textbf{15-step ahead} \\
\midrule
\multicolumn{4}{l}{\textit{Two-Layer LSTM Models}} \\
LSTM & 14.71 & 18.46 & 21.95 \\
SignatureLSTM-2-2 & 82.66 & 82.51 & 154.06 \\
SignatureLSTM-3-2 & 85.63 & 86.71 & 159.63 \\
SignatureLSTM-3-3 & 86.77 & 87.78 & 160.49 \\
SignatureLSTM-4-4 & 97.23 & 93.57 & 174.18 \\
\midrule
\multicolumn{4}{l}{\textit{Two-Layer GRU Models}} \\
GRU & 19.08 & 19.19 & 25.04 \\
SignatureGRU-2-2 & 41.74 & 47.77 & 50.31 \\
SignatureGRU-3-2 & 47.77 & 51.98 & 58.42 \\
SignatureGRU-3-3 & 43.89 & 46.83 & 53.00 \\
SignatureGRU-4-4 & 51.15 & 53.95 & 65.14 \\
\midrule
\multicolumn{4}{l}{\textit{Three-Layer Models}} \\
LSTM & 22.82 & 24.01 & 30.49 \\
SignatureLSTM-3-3-3 & 122.44 & 126.40 & 215.36 \\
GRU & 28.58 & 26.55 & 41.78 \\
SignatureGRU-3-3-3 & 66.96 & 62.16 & 78.30 \\
\midrule
\multicolumn{4}{l}{\textit{Flattened Output Models}} \\
LSTM & 18.31 & 17.70 & 25.38 \\
SignatureLSTM-3-3 & 128.54 & 87.42 & 162.89 \\
GRU & 18.78 & 20.80 & 28.87 \\
SignatureGRU-3-3 & 48.97 & 45.55 & 59.60 \\
\midrule
\multicolumn{4}{l}{\textit{Extended Signature Input Models}} \\
SignatureLSTM-2-2 & 89.37 & 87.39 & 155.90 \\
SignatureGRU-2-2 & 46.29 & 47.30 & 52.02 \\
\bottomrule
\end{tabular}
}
\label{tab:volume_training_times}
\end{table}
These comprehensive results reveal several key insights about model architecture choices:
First, the effectiveness of LSTM versus GRU architectures is highly task-dependent. While LSTMs excel at volatility prediction, GRUs prove superior for volume forecasting. This underscores the importance of maintaining a diverse model portfolio rather than seeking a one-size-fits-all solution. Second, the flattening strategy shows task-specific benefits. For volatility prediction, flattened architectures perform competitively with their sequential counterparts, particularly in longer horizons. However, for volume prediction, flattening generally leads to decreased performance, suggesting that maintaining temporal structure is more critical for this task. Third, signature-based modifications consistently improve upon the best-performing base architecture for each task. When LSTMs perform better (volatility), SignatureLSTM variants achieve the highest scores. Similarly, when GRUs are superior (volume), SignatureGRU models lead the performance metrics. This pattern suggests that signature-based gating mechanisms effectively enhance the underlying strengths of each architecture rather than fundamentally altering their characteristics. The optimal signature depth appears to be task-specific, with depth-3 models generally providing the best balance of performance and computational cost. Increasing to depth-4 shows diminishing returns relative to the additional computational overhead, while depth-2 models sometimes lack sufficient expressiveness for complex patterns. These findings highlight the adaptability of signature-based modifications across different architectural choices and prediction tasks, while also emphasizing the importance of careful model selection based on specific application requirements and computational constraints.

\section{Conclusion}
This paper introduces a novel approach to enhancing recurrent neural networks through the integration of path signatures into their gating mechanisms. By modifying the forget gate in LSTMs and the reset gate in GRUs, our method leverages the rich geometric information captured by path signatures to improve the networks' ability to process sequential data.
Our comprehensive empirical evaluation across two distinct forecasting tasks reveals several significant findings. The signature-based modifications consistently improve performance over traditional architectures, with particularly notable gains in longer-horizon predictions. The effectiveness of these enhancements varies between LSTM and GRU architectures in a task-dependent manner, demonstrating the flexibility of our approach in augmenting different base architectures. The computational overhead introduced by signature calculations, while significant, is offset by meaningful improvements in predictive performance and model stability. The SignatureGRU variant, in particular, offers an attractive balance between computational efficiency and prediction accuracy, making it particularly suitable for practical applications. These results open several promising directions for future research. The success of signature-based gating suggests potential applications in other neural architectures beyond traditional RNNs. Additionally, investigating more efficient methods for signature computation could further enhance the practical applicability of these models. The task-dependent nature of performance improvements also warrants further study into the relationship between data characteristics and optimal model configuration.
Our work contributes to the growing body of research combining classical mathematical techniques with modern deep learning architectures. The consistent improvements achieved across different tasks and architectures demonstrate the potential of path signatures as a powerful tool for enhancing sequential data processing in neural networks.

\bibliographystyle{IEEEtran}
\bibliography{bib}
\end{document}